\documentclass[letterpaper, 10 pt, conference]{ieeeconf}  

\IEEEoverridecommandlockouts                              

\usepackage{cite}
\usepackage{amsmath,amssymb,amsfonts}
\usepackage{algorithmic}
\usepackage{algorithm}
\usepackage{graphicx}
\usepackage[dvipsnames]{xcolor}
\usepackage{booktabs} 
\usepackage[]{units} 
\usepackage{subcaption}
\usepackage[short,c2,nocomma]{optidef}
\usepackage{hyperref} 
\hypersetup{
    colorlinks,
    linkcolor={blue!60!black},
    citecolor={blue!60!black},
    urlcolor={black}
}
\definecolor{dgreen}{RGB}{50, 180, 0}
\usepackage{fancyhdr}
 \usepackage[pscoord]{eso-pic}
\usepackage{lipsum} 

\fancypagestyle{firstpage}{%
    \fancyhf{} 
    \fancyfoot[L]{\parbox{\textwidth}{\centering \small  
         Accepted for publication in 2025 IEEE Conference on Robotics and Automation (ICRA).\\
         © 2025 IEEE. Personal use of this material is permitted. Permission from IEEE must be obtained for all other uses, in any current or future media, including reprinting/republishing this material advertising or promotional purposes, creating new collective works, for resale or redistribution to servers or lists, or reuse of any copyrighted component of this work in other works.  
    }}
}
\title{\LARGE \bf
Actor-Critic Cooperative Compensation to Model Predictive Control for Off-Road Autonomous Vehicles Under Unknown Dynamics
}

\author{Prakhar Gupta$^{1}$, Jonathon M. Smereka$^{2}$ and Yunyi Jia$^{1}$
\thanks{$^{1}$Prakhar Gupta ({\tt\footnotesize prakhag@clemson.edu}) and Yunyi Jia ({\tt\footnotesize yunyij@clemson.edu}) are with the Department of Automotive Engineering, Clemson University, Greenville, SC 29607, USA.}%
\thanks{$^{2}$Jonathon M. Smereka ({\tt\footnotesize{jonathon.m.smereka.civ@army.mil}}) is with the Ground Vehicle Systems Center, U.S. Army Combat Capabilities Development Command, Warren, MI 48397 USA.}%
\thanks{Acknowledgment: This work was supported by Clemson University’s Virtual Prototyping of Autonomy Enabled Ground Systems (VIPR-GS), a US Army Center of Excellence for modeling and simulation of ground vehicles, under Cooperative Agreement W56HZV-21-2-0001 with the US Army DEVCOM Ground Vehicle Systems Center (GVSC).}
\thanks{DISTRIBUTION STATEMENT A. Approved for public release; distribution is unlimited. OPSEC \#8979}%
}

\begin{document}

\pagestyle{empty}

\maketitle
\thispagestyle{firstpage}

\begin{abstract}
This study presents an Actor-Critic Cooperative Compensated Model Predictive Controller (\texttt{AC\textsuperscript{3}MPC}) designed to address unknown system dynamics. To avoid the difficulty of modeling highly complex dynamics and ensuring real-time control feasibility and performance, this work uses deep reinforcement learning with a model predictive controller in a cooperative framework to handle unknown dynamics. The model-based controller takes on the primary role as both controllers are provided with predictive information about the other. This improves tracking performance and retention of inherent robustness of the model predictive controller. We evaluate this framework for off-road autonomous driving on unknown deformable terrains that represent sandy deformable soil, sandy and rocky soil, and cohesive clay-like deformable soil.
Our findings demonstrate that our controller statistically outperforms standalone model-based and learning-based controllers by upto 29.2$\%$ and 10.2$\%$. This framework generalized well over varied and previously unseen terrain characteristics to track longitudinal reference speeds with lower errors.
Furthermore, this required significantly less training data compared to purely learning-based controller, while delivering better performance even when under-trained.
\end{abstract}

\def\mycopyrightnotice{
    {\footnotesize This work has been submitted to the IEEE for possible publication. Copyright may be transferred without notice, after which this version may no longer be accessible.\hfill}
}


\section{INTRODUCTION}

Aggressive off-road autonomous driving for vehicle platforms presents the challenge of real-time capable controller design that can handle the complexity of the system dynamics.
This complexity is owed to vehicle inertia and tires' interaction with the uneven and deformable terrain. It is difficult to model a high-fidelity plant model for the controller, and to even formulate a real-time controller for the highly non-linear plant.

Model-based controllers offer theoretical guarantees but computationally efficient vanilla schemes using simpler models as in \cite{KinematicBicycleModel2015}, fail to capture off-road conditions. This performance loss due to model mismatches is discussed in this paper and in \cite{gupta2024reinforcementlearningcompensatedmodel}. To some extent, dynamics uncertainties have been handled with robust \cite{bemporad2007robust} and stochastic optimal control schemes \cite{Mesbah, saltik2018outlook}. These formulations are complex and computationally demanding for real-time off-road applications. Adaptive methods as in \cite{bujarbaruah2019adaptive, rosolia2017learning, sinha2022adaptive} generally struggle with assumptions of uncertainty bounds. Purely learning-based approaches such as \cite{lillicrap2015continuous, schulman2017proximal} can adapt to specific scenarios very well but lack inherent safety assurances and require substantial data unless supplemented with techniques like barrier functions \cite{safeRL}.

Hybrid approaches that leverage the best of both worlds have the potential to improve performance and safety. Learning-based optimal control methods such as those using Gaussian processes \cite{hewing2020learning, fisac2018general} face scalability issues, while sampling-based control \cite{williams2017model, bootsRL, 10190183} generally fails to address state constraints effectively. Techniques that primarily rely on learnt networks during deployment \cite{bojarski2016end, salzmann2023real, spielberg2021neural, sacks2023deep} often require extensive data and may lack confidence in out-of-distribution scenarios. Most of these address on-road driving or small quad-rotors applications, leaving a gap in applying hybrid reinforcement learning controls to vehicles on highly deformable terrains. Implicit cloning of model-based control behavior can be achieved though methods like imitation learning \cite{kim2024reinforcement}, and Residual Policy Learning (RPL) to further improve the imitated controller with more data has been exploited in \cite{silver2018residual}, \cite{kerbel2023adaptive}.

\begin{figure*}
    \centering
    \includegraphics[width=\linewidth]{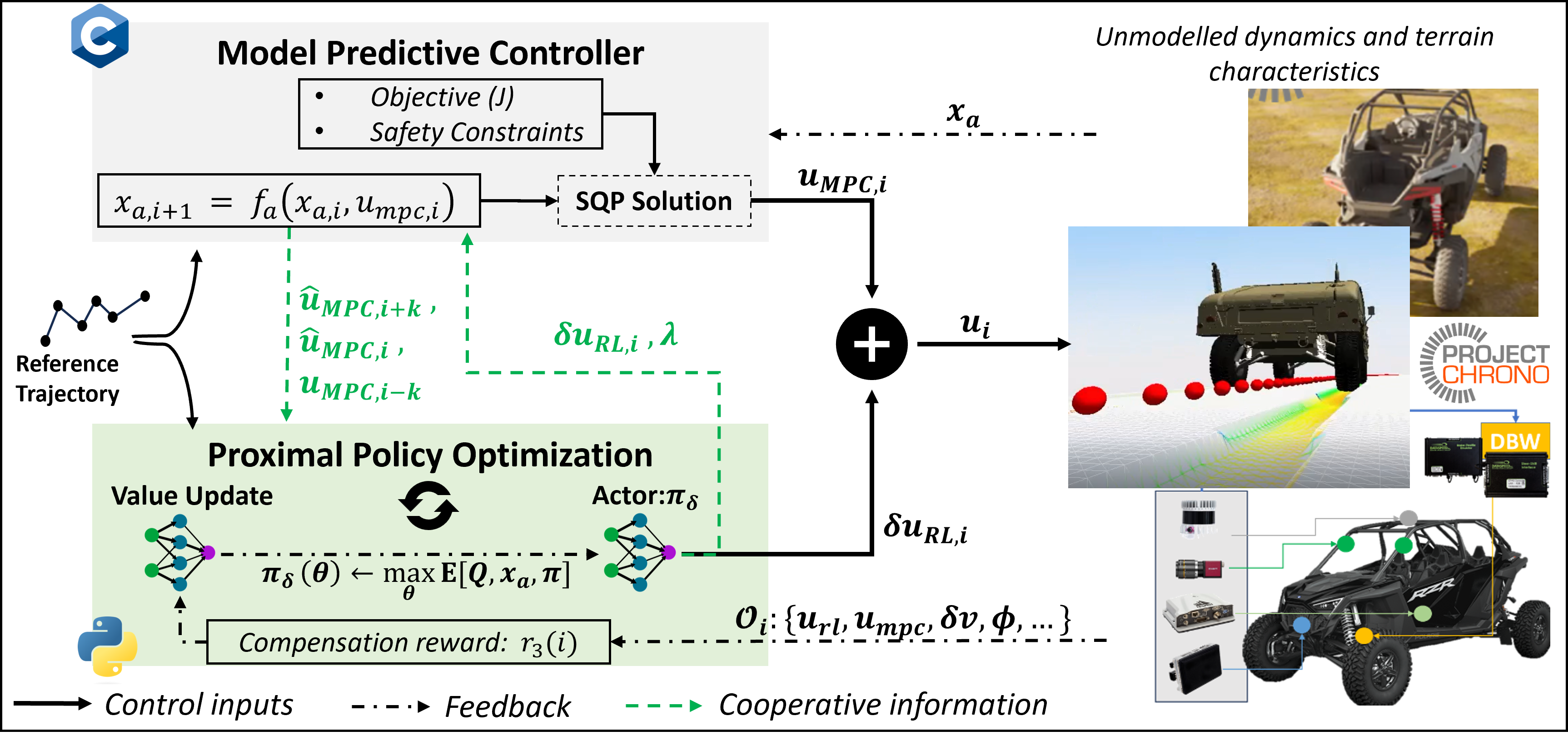}
    \caption{\texttt{AC\textsuperscript{3}MPC} controller training and simulation framework utilizes optimal control and learning paradigms. Cooperative and predictive information is exchanged to build anticipation of compensation and allows MPC to dictate control.}
    \label{fig:architecture}
\end{figure*}

Other efforts that specifically combine model predictive control (MPC) and reinforcement learning (RL) paradigms require prior parameterization or knowledge of model structure. These include tuning the former's parameters through RL \cite{zanon2020safe}, updating cost functions \cite{bhardwaj2020blending, arroyo2022reinforced}, embedding a differentiable controller into a network \cite{scarramuzza}.
Compensation approaches that require no knowledge of non-linear model structure are adopted in \cite{gupta2024reinforcementlearningcompensatedmodel, delft} where the agent learns to manipulate the primary controller's inputs for unknown residual dynamics. However, they do not consider the effect of compensation on the MPC predictions that are unaware of this manipulation.
This lack of synergy can lead to loss of tracking performance, receding horizon MPC's inherent robustness or feasibility. The framework proposed in this work (Fig.~\ref{fig:architecture}) seeks to address these challenges by developing a more data-efficient and cooperative hybrid control framework for high-speed off-road driving on unknown, deformable terrains.

\subsection{Contribution}
The contributions of this work are as follows.
\begin{itemize}
    \item We developed a cooperative parallel compensation architecture \texttt{AC\textsuperscript{3}MPC} to improve longitudinal performance of a model predictive controller  for unknown complex dynamics using proximal policy optimization.
    \item To improve tracking and retention of the predictive controller's inherent robustness, we built cooperation and awareness of the compensatory learning-based controller into the framework.
    \item The framework only needs to learn a cooperative behavior, requiring less data than a pure learning technique. Even when under-trained, our framework outperforms both standalone learnt and model-based controllers.
    \item The agent learns to intervene only in the operating regions where primary controller cannot achieve target tracking, improving utilization of the control space and smoothness of the control signal.
\end{itemize}

\section{COOPERATIVE COMPENSATION FRAMEWORK} \label{sec:fw}
\subsection{Baseline Controllers}
We establish three controllers from \cite{gupta2024reinforcementlearningcompensatedmodel} that our framework is based on: an \texttt{MPC} controller with no understanding of model mismatch, a purely reinforcement learning Actor-Critic (\texttt{AC}) controller, and the compensated controller \texttt{AC\textsuperscript{2}MPC} which lacks cooperation.

The \texttt{MPC} formulation represents the vanilla scheme using a non-linear kinematic bicycle model from \cite{KinematicBicycleModel2015} as its system dynamics constraint $ f(x_i, u_{mpc,i})$. Vehicle actuation and state constraints are made to match a real vehicle platform. For the \texttt{AC} controller using proximal policy optimization (PPO) \cite{schulman2017proximal}, the reward function is chosen as in Eq. \eqref{eq:r1}. 
\begin{subequations} \label{eq:r1}
    \begin{align}
        r_1(i) &= - W_{11} \cdot \frac{\left| v_{err} \right|}{N_e} - W_{12} \cdot \frac{\sigma_{a_{rl}}}{N_{\sigma}} - W_{13} \cdot p_{13}\\
            & \text{s.t. } p_{13}=\begin{cases}
            1, & \text{if $v < 0$}\\
            0, & \text{otherwise}.
          \end{cases}
    \end{align}
\end{subequations}
Here, we penalize the speed tracking error  $v_{err}$ and negative velocities through $p_{13}$. Standard deviation $\sigma_{a_{rl}}$ over the observed action history span of $h_a$ is penalized to encourage smooth control. These components are all normalized to unity with the help of appropriate normalizer such as $N_{\sigma}$ to improve learning. The corresponding weight terms $W_{1k}$ are non-negative real numbers. 

The compensated controller \texttt{AC\textsuperscript{2}MPC} includes the same MPC formulation as the baseline MPC and is compensated using an actor-critic which observes the past MPC actions, past agent actions, and past tracking errors. The reward function is shown in Eq. \eqref{eq:r2}. Since the model-based controller is expected to track small reference velocities well, $p_{23}$ is used to discourage unnecessary agent compensation at low speeds. We refer to this controller as the unaware compensation controller because the primary controller generates a control input unaware of the compensatory controller. This means that the cooperative information flow depicted in Fig.~\ref{fig:architecture} is missing in this controller. The compensated inputs could thus be thought of as an external unobserved disturbance for the MPC that it may adjust to, in order to perform speed tracking. This could jeopardize the predictive controller's inherent robustness and recursive constraint feasibility because the predictions over the horizon may not hold true once the compensation input is added.
\begin{subequations} \label{eq:r2}
    \begin{align}
        r_2(i) & = W_{21} \cdot \frac{1}{1+ \left| v_{err} \right| } - W_{22} \cdot \frac{\sigma_{a_{rl}}}{N_{\sigma}} - p_{13} - W_{23} \cdot p_{23} \\
            & \text{s.t. } p_{23}=\begin{cases}
            1, & \text{if $a_{rl} > 0 \ \& \ v < v_{threshold}$}\\
            0, & \text{otherwise}.
          \end{cases}
    \end{align}
\end{subequations}

\subsection{Cooperative Compensation Controller}
In the \texttt{AC\textsuperscript{3}MPC} controller, we add awareness and anticipation for compensation by augmenting the MPC system dynamics to be $f_a(x_i, u_i)$ as in Eq. \eqref{eq:AugmentedDyn}. 
\begin{subequations} \label{eq:AugmentedDyn}
    \begin{align}
        \Dot{s}_x &= v \cos{(\phi + \beta)} \\
        \Dot{s}_y &= v \sin{(\phi + \beta)} \\
        \Dot{\phi} &= \frac{v}{L} \tan{(\beta)} \\
        \Dot{\delta} &= \omega \\
        \Dot{v} &= a_{mpc} - a_{rl} \\
        \beta &= {\tan}^{-1}(\frac{L_r}{L} \cdot \tan{\delta}) \\
        \Dot{a_{rl}} &= \lambda 
    \end{align}
\end{subequations}
The augmented state vector presented includes positions in X, Y directions, yaw angle, steering angle, speed, body slip angle and additionally, the compensated input from the current control step respectively. This augmentation is much like a disturbance rejection formulation of an MPC with disturbance observer, but in this case, we treat it as a desirable disturbance. This allows the MPC to predict and optimize for the agent's actions to yield improved performance. This also increases the chance of constraint feasibility along the horizon. The bounds for $a_{rl}$ are tighter than those for $a_{mpc}$, which is allowed to span the entire available throttle actuation range as a primary controller. These reduced action bounds for the agent are justified through the reasoning that we intend to learn a corrective input above the primary control input. Limiting this correction range allows improving data efficiency while decreasing tracking errors.
The rate of change of $a_{rl}$, $\lambda$ is chosen to be a small value so that the expected compensation by the MPC does not violate bounds, but still allows understanding of the compensation effect over the horizon. The new optimization problem with the augmented states and dynamics is given by Eq. \eqref{eq:ll-mpcrlc}.
\begin{mini!}
    {
        \mathbf{u_{mpc}}, \, \mathbf{x}}{
        \left\Vert {\mathbf{x}_N}_a - {\chi_{N}} \right\Vert^2_{P_a} + \ldots
    }{\label{eq:ll-mpcrlc}}{}
    \breakObjective{\sum_{i=0}^{N-1} \left( 
            \left\Vert {\mathbf{x}_i}_a - {\chi_{i}} \right\Vert^2_{Q_a} + \left\Vert {\mathbf{u_{mpc}}_{i}} - \mu_{i} \right\Vert^2_R
        \right)}{}
    \addConstraint{
        {x_{i+1}}_a = f_a({x_i}_a, u_{mpc,i})
    }{}
    \addConstraint{\label{eq:llmpc_outputs}
        z_i = a_l = \frac{v_i ^ 2}{L} \tan(\delta_i)
    }{}
    \addConstraint{\label{eq:llmpc_c1}
        0 \leq v_i
    }{}
    \addConstraint{\label{eq:llmpc_c2}
        \underline{a}_l \leq a_l \leq \overline{a}_l
    }{}
    \addConstraint{\label{eq:llmpc_c3}
        \underline{a_{mpc}} \leq a_i \leq \overline{a_{mpc}}
    }{}
    \addConstraint{\label{eq:llmpc_c4}
        \underline{\delta} \leq \delta_i \leq \overline{\delta}
    }{}
    \addConstraint{\label{eq:llmpc_c5}
        \underline{\omega} \leq \omega_i \leq \overline{\omega}
    }{}
\end{mini!}
where,
$ x_{i,a} $ is the vehicle state vector at any given time $i$; $ u_{i} = \left[ a, \omega \right] $ is the control input vector of forward acceleration and rate of change in steering angle $\delta$; $z_i$ is the algebraic state approximating lateral acceleration; $\chi_i$ and $\mu_i$ are the respective reference states and controls at each control stage $i$ corresponding to the trajectory planned ahead in time; $ Q_a \succeq 0 $, $P_a \succeq 0 $, $ R \succ 0 $ are the state penalty, terminal state penalty, input penalty matrices respectively. Equations (\ref{eq:llmpc_outputs})-(\ref{eq:llmpc_c5}) are the constraints imposed on the states and control inputs.

Further, the learning agent also observes the primary controller's next predicted inputs ($\hat{u}_{mpc,i+1}$) along with current inputs to learn the primary policy's temporal aspect. This cooperative information exchange at every control step is depicted in Fig.~\ref{fig:architecture}. For training the network, the length of action history $h_a$ that is observed by the agent is rendered much smaller compared to the previous unaware framework, reducing the size of observation space $\mathcal{O}$. 
The reward function for this controller is $r_3$ which penalizes tracking errors with a weight $W_{31}$. It employs the boolean $p_c$ to penalize unnecessary compensation inputs. This way, compensation is promoted only when it improves tracking performance and the primary control input is under the control ceiling.
$p_{33}$ is used to penalize agent inputs that lead to a violation of the vehicle's actuation limits as in Eq. \eqref{eq:r3}. The algorithmic flow for our framework is given in Algorithm \ref{alg:mpcrlc}, where the cooperative aspects are in green.
\begin{subequations} \label{eq:r3}
    \begin{align}
        \begin{split}
            r_3(i) & = - p_{c} \cdot W_{31} \cdot \frac{\left| v_{err} \right|}{N_e} - (1-p_{c}) \cdot W_{32} \cdot a_{rl}^2 \\
            &  \quad \quad \quad \quad \quad \quad \quad - W_{33} \cdot \frac{p_{33}}{N_{violate}} \\
        \end{split}\\
        & \text{s.t. } p_{c} = \begin{cases}
            1, & \text{if $a_{min} + \epsilon <  a_{mpc} < a_{max} - \epsilon $}\\
            0, & \text{otherwise}.
          \end{cases}\\
        & \text{s.t. } p_{33} = \begin{cases}
            a_{mpc} + a_{rl} - a_{max}, & \text{if $(a_{mpc} + a_{rl}) > 1$}\\
            a_{min} - a_{mpc} - a_{rl}, & \text{if $(a_{mpc} + a_{rl}) < -1$}\\
            0, & \text{otherwise}.
          \end{cases}
    \end{align}
\end{subequations}
%
\begin{algorithm}[h]
\caption{\texttt{AC\textsuperscript{3}MPC}: Parallel cooperative compensation}\label{alg:mpcrlc}
\begin{algorithmic}

\STATE \COMMENT{for each control step}
\STATE \text{\textbf{Input}: $x_{actual,i-1}$, $u_{i-1}$}
\STATE $\text{Simulate environment for n timesteps with}$ $u_{i-1}$ 
\STATE  \hspace{0.5cm} $x_{actual,i} \gets f_{actual}(x_{i-1},u_{i-1},T_s) $
\STATE $\text{Generate observations and feedback}$ 
\STATE  \hspace{0.5cm} $x_{actual,i}, s_{obs} \gets sim()  $
\STATE     $\text{If training PPO:}$ 

      \hspace{0.5cm} (if sample $\geq n_{steps}$):
      
         \hspace{0.9cm} $\theta_{i+1} \gets \arg\max_{\theta} E[\mathcal{L}(\mathcal{O}_i, u_{rl}, \theta_i, \theta)] $
        
        \hspace{0.9cm} $\mathcal{L} = \min( \frac{\pi_{\theta}}{\pi_{\theta_{i}}} \cdot \mathcal{A}_i(\mathcal{O},u_{rl}), (1 \pm \epsilon) \cdot \mathcal{A}_i(\mathcal{O},u_{rl}))  $ 

\STATE $\text{Get agent action}$ 
\STATE  \hspace{0.5cm} $u_{rl, i} \gets \pi_{\theta}({x_{i}}_a, \textcolor{dgreen}{u_{mpc, i-k}, \hat{u}_{mpc, i:i+k}})$

\STATE $ \text{Solve MPC}  $     
\STATE       \hspace{0.5cm} \texttt{\small Exchange predictive information}  

\hspace{0.5cm} $u_{mpc, i}, \hat{u}_{mpc, i+k} \gets \pi_{mpc}(\rho_{i},\mu_i, \textcolor{dgreen}{\lambda}, x_{i, a}) $

\STATE      $\text{Compensate}$ 
\STATE \hspace{0.5cm} $u_{i} \gets u_{mpc, i}+ u_{rl, i} $

\STATE \hspace{0.5cm}\textbf{return}  $u_i$
\STATE \text{- - - - - - - - - - - - - - - - - - - - - - - - - - -  - - - - - - -}
\STATE \text{\footnotesize{Legend: \textcolor{dgreen}{ -- Cooperative information}}}

\end{algorithmic}
\end{algorithm}

\begin{figure*}
    \centering
    \includegraphics[width=0.97\textwidth]{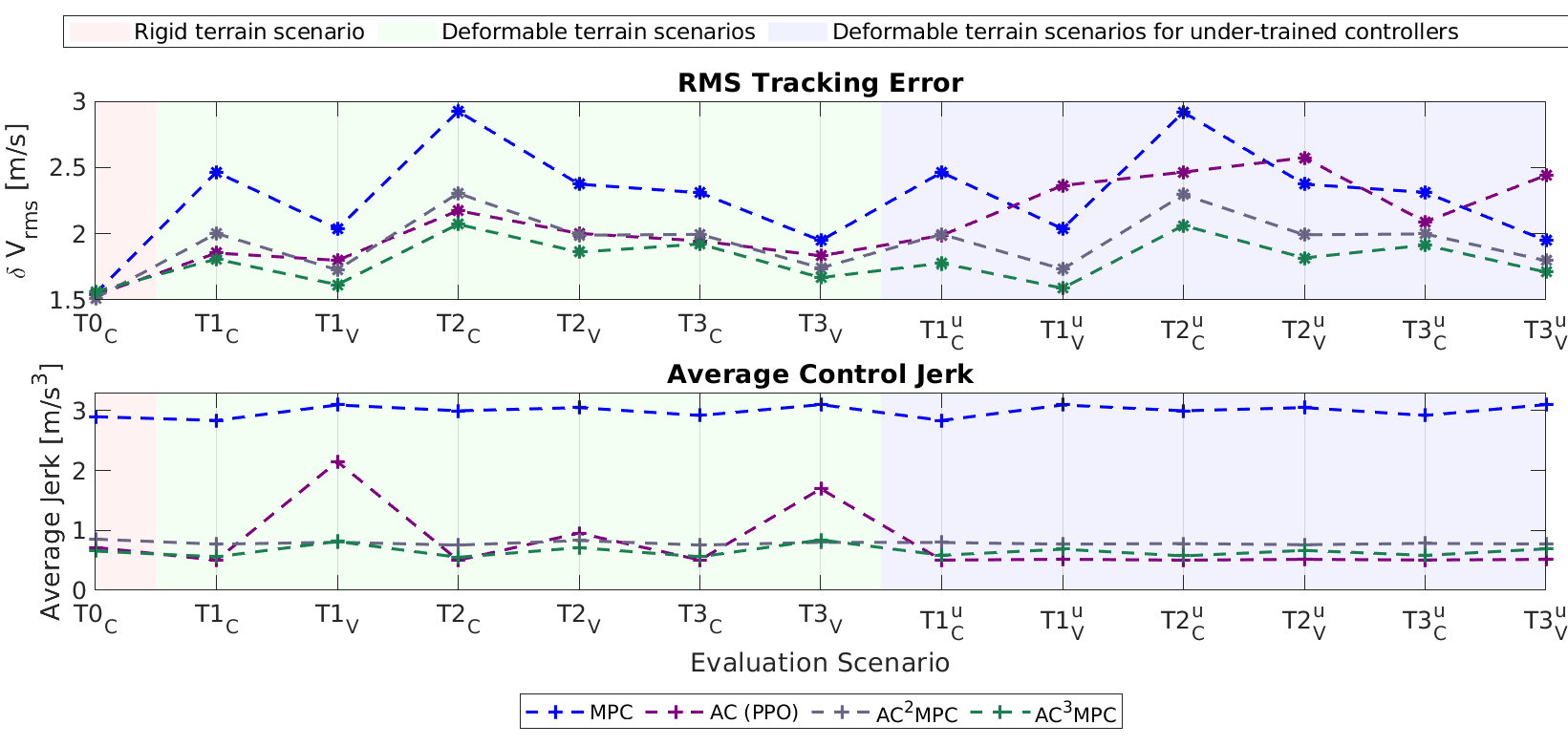}
    \caption{Performance of all controllers is plotted for 13 evaluation scenarios: Driving on rigid terrain ($T0_r$), driving on deformable terrains ($T1_r$,$T2_r$,$T3_r$), and driving on the same deformable terrains with under-trained controllers ($T1_r^u$, $T2_r^u$, $T3_r^u$). RMS tracking error and smoothness measure is plotted for all these scenarios and our framework has lowest errors and jerks in general, even when under-trained.}
    \label{fig:compare}
\end{figure*}

\section{SIMULATION RESULTS AND DISCUSSION} \label{Results}
The four controllers described in section \ref{sec:fw} are implemented using the \texttt{acados} \cite{acados}, \texttt{stable-baselines3} \cite{stable-baselines3} and \texttt{gymnasium} \cite{gym} libraries using a single \texttt{RTX A6000} GPU. Simulations are carried out in \texttt{Project Chrono} \cite{Chrono2016} to train these controllers over random speed profiles. To match reality, where training data is sparse and seldom available for more than one terrain, we train on just one terrain ($T1$), which is a moderately loose deformable terrain.
To evaluate generalizability to unseen terrains and comparison with baseline controllers, we consider driving over the deformable terrain $T1$, deformable terrain $T2$ (a sandy and rocky terrain with higher stiffness) and deformable terrain $T3$ (a clay-like and soft terrain). 
For the rest of this study, we label the evaluation scenarios as $Tn_{r}^u$, where $n$ is the terrain number, $r$ refers to the reference type which can be Constant or Varying and $u$ depicts under-trained scenarios.
\subsection{Tracking Performance}
%
Fig.~\ref{fig:compare} plots and compares the key performance metrics for the four controllers we have discussed above. First, we examine scenario $T0_C$ which considers a rigid terrain with low model mismatch. With this, we establish that the \texttt{MPC} controller is fairly formulated to perform well enough on rigid terrains but tracks poorly in off-road scenarios when the model-mismatch is high as in $T1_C$. Having no understanding of the model-mismatch, it consistently performs poorly in all off-road scenarios. The plant model for this controller cannot consider the unknown deformable terrain dynamics and thus there exists a problem that can be addressed with our learning augmentation.
\begin{figure}
    \centering
    \begin{subfigure}{0.97\linewidth}
    \includegraphics[width=\linewidth]{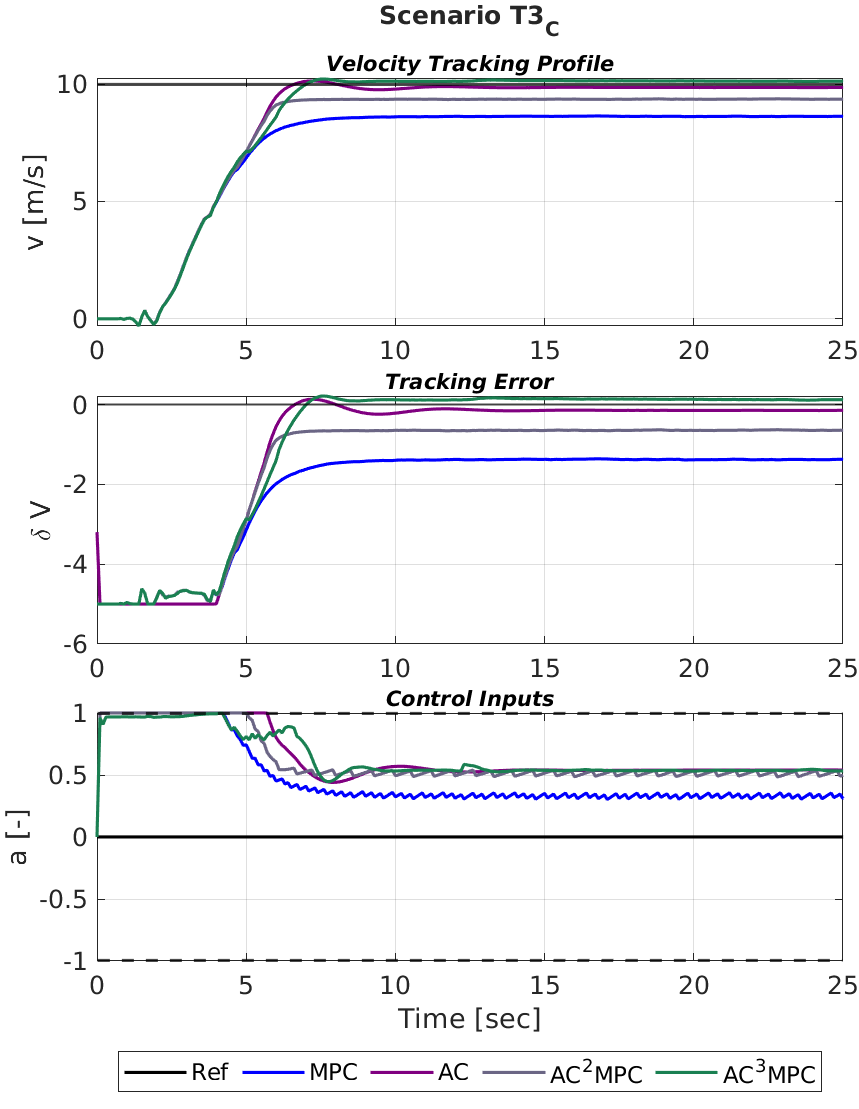}
    \end{subfigure}\vspace{1.0em}
    
    \begin{subfigure}{0.97\linewidth}
    \includegraphics[width=\linewidth]{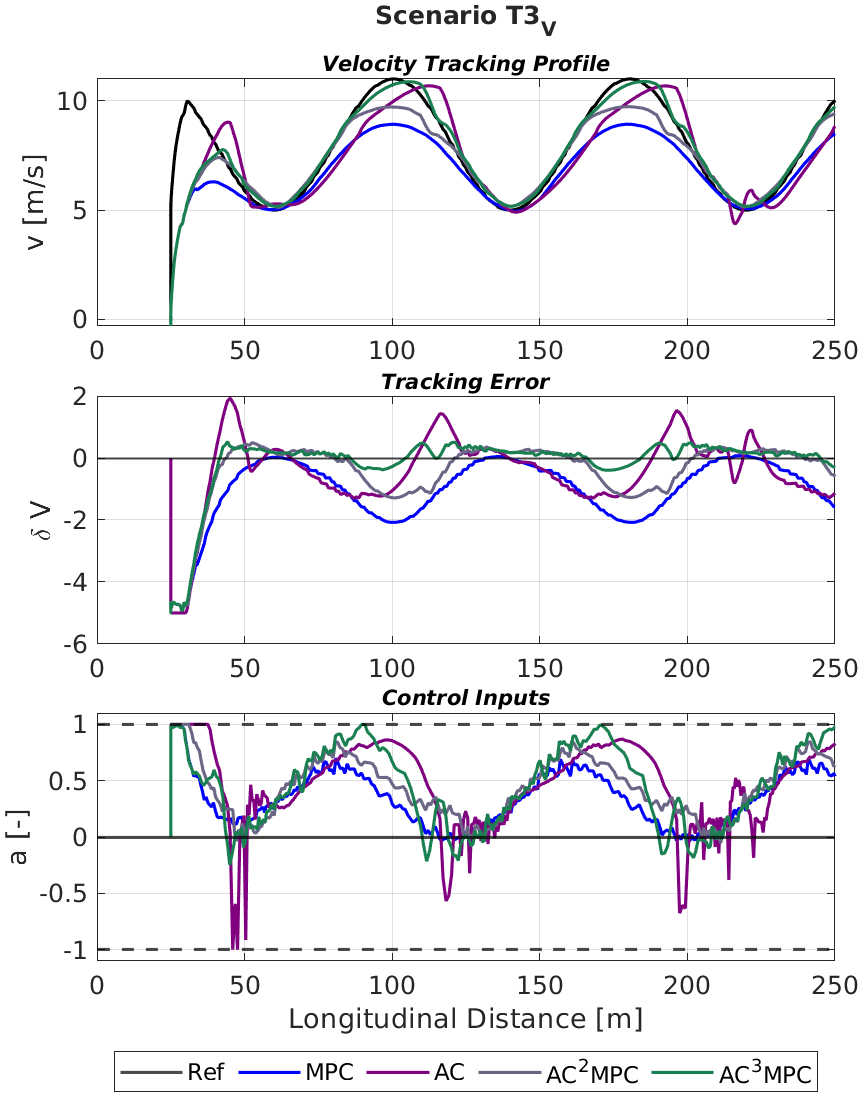}
    \end{subfigure}
    \caption{Controller comparison for velocity tracking on a previously unseen soft clay-like terrain $T3$ for constant and varying reference velocity scenarios $T3_C$ and $T3_V$ respectively.}
    \label{fig:s3AB}
\end{figure}

\texttt{AC} controller, shown in magenta in Fig.~\ref{fig:compare}, manages to track the reference velocities somewhat closely in all scenarios, but yields jerky control inputs when dealing with varying reference speeds. 
Despite a reward function design to penalize control jerks and encourage smoothness, large jitters are observed, as shown in Fig.~\ref{fig:s3AB} for scenarios $3_C$ and $3_V$.
The tracking delay, which is evident in the varying velocity profile plot of Fig.~\ref{fig:s3AB}, is owed to the fact that the learnt controller does not have an explicit prediction horizon like the model predictive controller. Since it relies on observations from the environment solely after the fact, this delay is unavoidable for this formulation. The tracking errors spike up significantly around \unit[50]{m}, \unit[120]{m} when acceleration inputs are required to be low and near zero. The tracking errors plotted in the figure are capped at \unit[5]{m/s} for clearer low error region visualization and match the reference saturation value for the controllers. 

The \texttt{AC\textsuperscript{2}MPC} that has also been trained over random speed profiles yields better results than the \texttt{MPC} but has little to no improvement over the \texttt{AC} controller. In contrast to results presented in \cite{gupta2024reinforcementlearningcompensatedmodel}, where the agent had the freedom to span the entire normalized control space $[-1.0, 1.0]$, the bounds of the agent's action space are lower such that $ a_{rl} \in [-0.33, 0.33]$ to match the bounds of the cooperative agent. At \unit[100]{m} in the second plot of Fig.~\ref{fig:s3AB}, it can be seen to undercut the target peak velocity, despite available control space. This is because the MPC component is unaware of the effects of the compensatory inputs on the system progress. It inaccurately interprets the compensation effects as arising from its own inputs and renders its own horizon predictions invalid. The optimizer then attempts to correct this by generating a lower control input for the next step and this unnecessary ramping down of primary control inputs prevents exploitation of the available control space.

\begin{figure}
    \centering
        \includegraphics[width=0.97\linewidth]{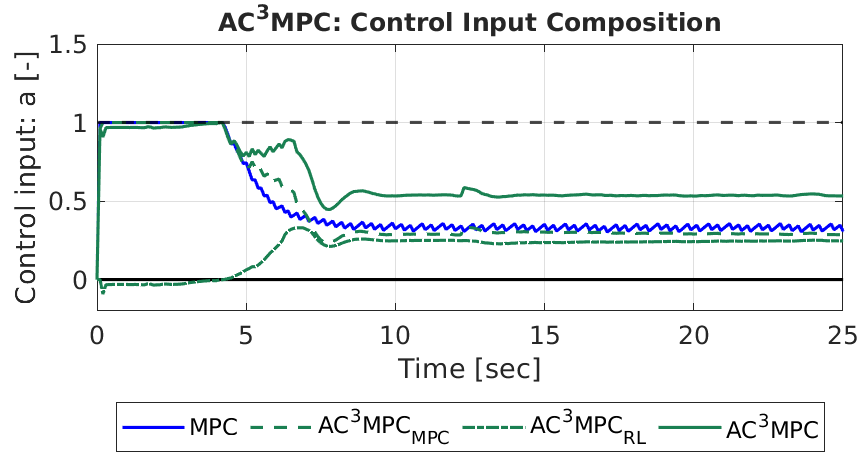}
        \caption{Control effort split for scenario $T3_C$: This figure shows the learnt compensation (AC\textsuperscript{3}MPC\textsubscript{RL}) that is added to the primary input (AC\textsuperscript{3}MPC\textsubscript{MPC}) while maintaining smooth control.}
    \label{fig:split}
\end{figure}

On the other hand, the \texttt{AC\textsuperscript{3}MPC} controller has consistently lower RMS tracking errors than all the others as tabulated in Table \ref{tab:results}. The predictive component lends the least amount of delays in tracking of the varying references here. In the first plot of Fig.~\ref{fig:s3AB}, it is seen that this controller tracks the constant target without any overshoot. At \unit[100]{m} in the second plot of Fig.~\ref{fig:s3AB}, this controller tracks the peaks of the varying profile well. This is because the optimization component understands the contribution of the learnt component to the overall progress in the simulation. This also allows a better exploitation of the control space than other controllers. This controller outperforms \texttt{AC\textsuperscript{2}MPC}'s error tracking by $3-10\%$ across all scenarios.
A closer look at the control composition for scenario $T3_C$ is provided in Fig.~\ref{fig:split}. The MPC component of the cooperative controller retains a behavior that is very close to the standalone model-based controller. This allows the agent to compensate without unnecessary counter-action and renders the model-based component's control to be smoother than the standalone controller. 
In terms of control constraint obedience, our framework attempts to stay within control bounds and attempts feasible control generation. We observe that the compensation input learns to ramp up only after the MPC inputs start dropping below the control ceiling without achieving a high enough velocity. Even though the generated control can be saturated to match vehicle actuation limits, it is advantageous for tracking that the framework implicitly understands this requirement.

\begin{table}
    \centering
    \caption{Statistics for all deformable terrain scenarios and controllers: RMS tracking error and average jerk values for one complete simulation for each scenario.}
    \begin{tabular}{r|l|ccc}
        \toprule
    	 Scenario & Controller	&	$\delta V_{rms}$ [m/s]  &	Avg. Jerk [m/s$^3$]	\\
        \midrule	
        \midrule									

        Scenario $T1_C$ & MPC	&	2.465  &	2.835	 \\
                    & AC	&	1.854 & \textbf{0.503} \\
& AC\textsuperscript{3}MPC	&	\textbf{1.808} & 0.559 \\
        \midrule	
        
        Scenario $T1_V$ & MPC	&	2.037  &	3.093	 \\
                     & AC	&	1.799 & 2.147 \\
 & AC\textsuperscript{3}MPC	&	\textbf{1.615} & \textbf{0.815} \\
        \midrule	
        



        Scenario $T2_C$ & MPC	&	2.924  & 2.996 \\
                    & AC	&	2.174 & \textbf{0.502}\\
 & AC\textsuperscript{3}MPC	&	\textbf{2.071 }& 0.551 \\
        \midrule									
        Scenario $T2_V$ & MPC	&	2.374  & 3.049 \\
                     & AC	&	2.002 & 0.954 \\
 & AC\textsuperscript{3}MPC	&	\textbf{1.861} & \textbf{0.711} \\     
        \midrule						


        Scenario $T3_C$ & MPC	&	2.309  & 2.918	 \\
                     & AC	&	1.946 & \textbf{0.506} \\
 & AC\textsuperscript{3}MPC	&	\textbf{1.922} & 0.560 \\
        \midrule									
        Scenario $T3_V$ & MPC	&	1.948  & 3.099\\
                     & AC	&	1.833 & 1.694 \\
 & AC\textsuperscript{3}MPC	&	\textbf{{1.668}} & \textbf{0.841} \\   


        \midrule									
         \textbf{Average}  & MPC	&	2.3428  & 2.9983\\
          (over all scenarios)           & AC	&	1.9346 & 1.0510 \\
 & AC\textsuperscript{3}MPC	&	\textbf{1.8241} & \textbf{0.6728} \\   
 & AC\textsuperscript{2}MPC	&  1.9681  &  0.7855 \\

        \bottomrule
    \end{tabular}
    \label{tab:results}
\end{table}

\subsection{Data Requirements and Under-trained Behavior}
\begin{figure}[h] 
    \centering
    \begin{subfigure}{0.97\linewidth}
    \centering
    \includegraphics[width=\linewidth]{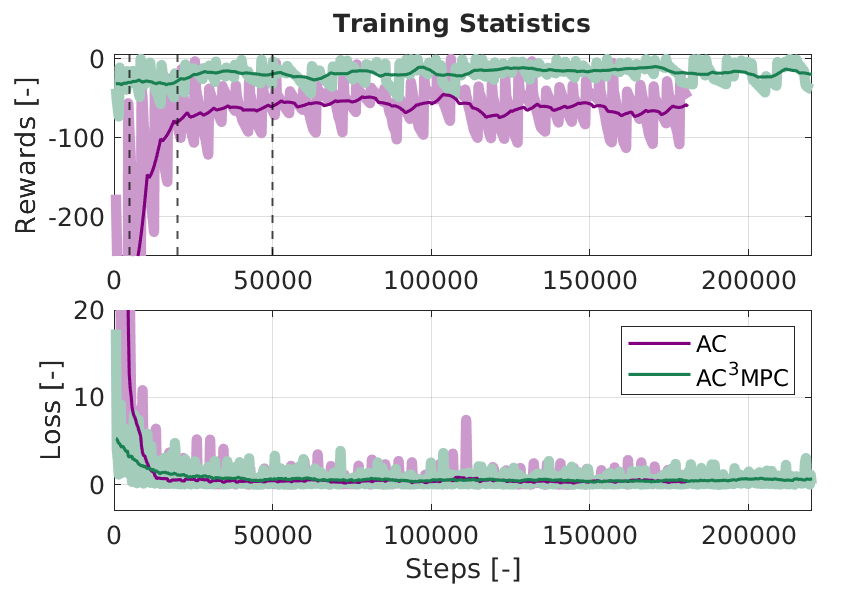}
    \caption{Rewards for AC, AC\textsuperscript{3}MPC are shown here. It is seen that the cooperative compensated controller starts at a high reward value already and converges within fewer steps.}
    \end{subfigure}
    \begin{subfigure}{0.97\linewidth}
    \includegraphics[width=\linewidth]{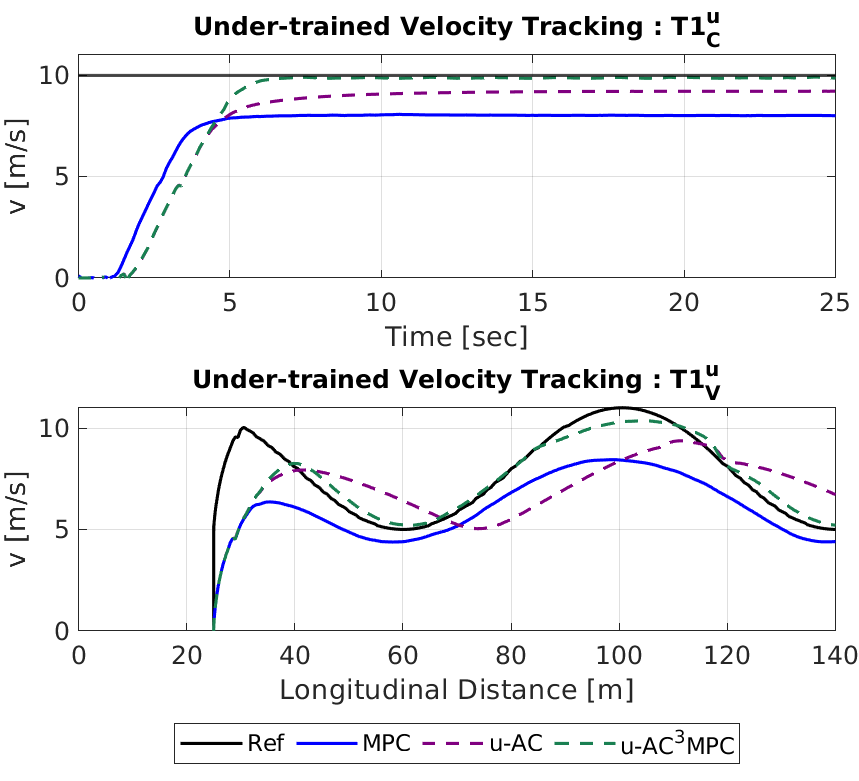}
    \caption{At only 5,000 steps of training, the performance of under-trained AC\textsuperscript{3}MPC (u-AC\textsuperscript{3}MPC) is already superior to under-trained AC (u-AC) and MPC.}
    \end{subfigure}
    \caption{Data requirements for compensated controller framework are lower. Performance for scenarios $T1_C^u$ and $T1_V^u$ is visualized here}
    \label{fig:data}
\end{figure}

Fig.~\ref{fig:data}(a) shows the reward and loss trends during training over randomized speed profiles. The reward for the cooperative controller training starts a small negative but quickly ramps up to near its saturation value within $20,000$ time-steps. The AC controller on the other hand, takes around $40,000$ time-steps to reach near its saturation value. This is because the this controller has to learn a policy from scratch, spanning the entire control actuation space for the vehicle. The cooperative controller would do better than even the residual policy learning paradigm, because instead of first learning to imitate the model-based controller and then learning data-driven improvements, it only needs to learn the cooperative compensation behavior. The tracking performance of the \texttt{AC\textsuperscript{3}MPC} throughout the training remained similar and promising, but the \texttt{AC} controller severely under-performs when under-trained. Moreover, it was seen from evaluations that \texttt{AC} tends to overfit more than the \texttt{AC\textsuperscript{3}MPC} as it is allowed to train longer.

To investigate under-trained performance, we extract control policies from each controller at only $5,000$ steps and re-evaluate all deformable terrain scenarios. These are represented by scenarios $T1_r^u, T2_r^u, T3_r^u$ in Fig.~\ref{fig:compare}. As an example, under-trained tracking of fixed and varying reference profiles on terrain $T1$ is plotted in Fig.~\ref{fig:data}(b). It is clear that the purely learning based controller is unusable when under-trained, but our framework performs very similar to when fully trained and thus is less data dependent and more generalized.

\section{CONCLUSION AND FUTURE WORK}
In this paper, we developed and evaluated a cooperative parallel compensation architecture to improve the longitudinal tracking performance of model-based controllers operating with model mismatch and unknown system dynamics. We specifically studied off-road driving on deformable terrains with a full-scaled vehicle platform.
Evaluation over multiple simulation scenarios revealed that this controller architecture can perform statistically better than standalone model-based and learning controllers by up to $29.16\%$ and $10.21\%$ respectively in terms of RMS tracking errors. The cooperative exchange of information introduces a synergy between the model-based and the learning controllers, and allows retaining robustness of the model-based controller's horizon predictions. We also observe improvement in control jerk and generalizability across previously unseen scenarios as compared to pure learning method.
This framework required lesser data to train on, which shows promise for physical platform deployment that is likely to experience out-of-distribution states.
Future work will include steering control learning and deployment on a real drive-by-wire vehicle platform (Polaris RZR).


\addtolength{\textheight}{-7cm}   


\bibliographystyle{ieeetr}
\bibliography{refs}

\end{document}